%% file: main_arxiv.tex
\documentclass[letterpaper,10pt,conference]{ieeeconf}
\IEEEoverridecommandlockouts
\usepackage[T1]{fontenc}
\usepackage{mathptmx}
\usepackage{amsmath,amssymb}
\usepackage{graphicx}
\usepackage{flafter}
\usepackage{placeins}
\usepackage{float}
\usepackage[caption=false,font=footnotesize]{subfig}
\usepackage{booktabs,array}
\usepackage{xcolor}
\usepackage{cite}
\usepackage{url}
\usepackage{tikz}
\usepackage[colorlinks=true, citecolor=blue, linkcolor=blue, urlcolor=blue]{hyperref}
\usetikzlibrary{arrows.meta,positioning}
\hypersetup{pdftitle={MoCaP: Mobile Code-as-Policies for Zero-Shot Mobile Manipulation from Passive Human Videos},pdfauthor={Yiyang Cai, Tatsuya Kamijo, Ninghao Lu, Zijin Hu, Sherry Yang}}

\title{\LARGE\bf MoCaP: Mobile Code-as-Policies for\\
Zero-Shot Mobile Manipulation from Passive Human Videos}
\author{Yiyang Cai$^{1,2,*}$, Tatsuya Kamijo$^{1,*}$, Ninghao Lu$^{3}$, Zijin Hu$^{1}$, and Sherry Yang$^{1}$%
\thanks{$^{*}$Equal contribution.}%
\thanks{$^{1}$New York University, New York, NY, USA.}%
\thanks{$^{2}$Fudan University, Shanghai, China. Work done while Yiyang Cai was visiting New York University.}%
\thanks{$^{3}$NYU Shanghai, Shanghai, China.}%
}
\begin{document}
\maketitle
\thispagestyle{empty}
\pagestyle{empty}
\begin{abstract}
Passive human videos of everyday activity offer robots access to a massive amount of human experience without requiring a demonstration for each task. However, using this experience requires recovering knowledge from unstructured video recordings and adapting it to a different body (i.e., the robot). To overcome these challenges, we present mobile code-as-policy (MoCaP), an agentic framework that turns passive human video recordings into guidance for an end-to-end mobile manipulation policy. MoCaP extracts spatial memory and cues of how people prepare to interact with objects, which inform navigation and the search for a manipulation-ready configuration. Guided by passive human experience and current observations, MoCaP uses a closed-loop code-as-policy to coordinate planning and tool use primitives to solve motions under the robot's geometry and kinematic constraints. 
On a real mobile manipulator, MoCaP achieves 80\% success on each of two mobile manipulation tasks in an indoor environment whereas existing approaches cannot succeed at the task. Video-derived spatial memory improves object-navigation success from 70\% to 100\%, and human approach prior raises the rate of finding a reachable base pose from 0\% to 100\% at an unfavorable start location.
Videos and code are available on the \href{https://mocap-7.github.io/MoCAP-page/}{project website}.
\end{abstract}
\input{figures/figure1}
\input{sections/introduction}
\input{sections/related_work}

\input{sections/method}
\input{sections/experiments}

\FloatBarrier
\input{sections/conclusion}
\section*{Acknowledgment}
We thank Kanad Patel for help with the YOR hardware setup and Billy Yan for helpful discussions.
\bibliographystyle{IEEEtran}
\bibliography{references}
\end{document}

%% file: figures/figure1.tex
\begin{figure}[!t]
\centering
\includegraphics[width=\linewidth]{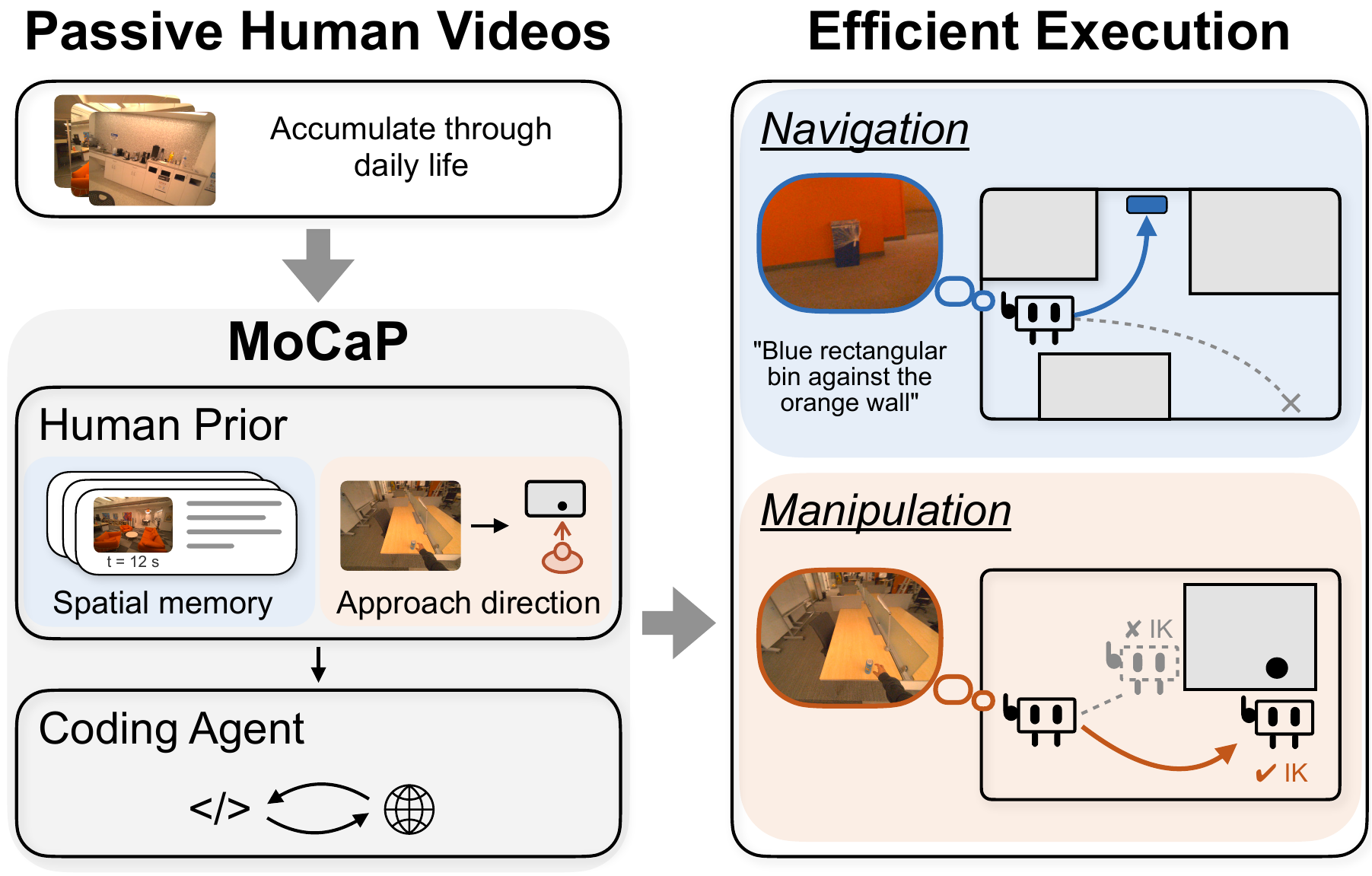}
\caption{\textbf{MoCaP turns passive human video into guidance for real-world mobile manipulation.} From videos recorded during daily life, MoCaP extracts a spatial memory and an approach direction. A coding agent uses them to find the target and reach a base pose with a feasible grasp. Grey dashed paths show execution without this prior.}
\label{fig:figure1}
\end{figure}

%% file: sections/introduction.tex
\section{Introduction}
Household robots should be able to assist different users in a wide range of homes, responding to everyday requests without requiring users to specify every action. Open-vocabulary mobile manipulation moves toward this goal by enabling robots to act on natural-language instructions involving diverse objects and environments~\cite{homerobot}. To fulfill a request such as ``bring the cup back,'' a robot must determine which objects and places are relevant, locate them, and decide how to act. Supporting this ability across homes and users motivates learning from experience beyond the robot's own interactions.

Passive recordings of human activity offer a potentially abundant source of such experience. Wearable cameras can capture how people interact with objects and their surroundings during everyday life, without requiring robot teleoperation or prescribed demonstrations~\cite{passiveactions}. We investigate how robots can draw on this accumulated human experience for open-vocabulary mobile manipulation. We consider the setting where a collection of human activity videos is passively recorded through people's daily lives capturing diverse activities that may be requested of a robot. Rather than reproducing a particular demonstration, the robot must use relevant experience from the collection to guide its actions in its current environment.

\input{figures/figure2}

However, passive recordings of egocentric human data do not directly provide the information a robot needs to act. Specifically, using passive human experience for robots presents three challenges.

 \textit{First, the recorded experience lacks explicit spatial structure.} A video supplies neither object coordinates nor labeled places and traversable connections. Relevant observations are distributed over time, with objects and regions reappearing from different viewpoints. Searching beyond the current view requires connecting objects to their spatial context.

\textit{Second, human and robot sensing and morphology differ.} Camera height and viewing angle change the visible scene, while different cameras introduce further differences in intrinsics, field of view, resolution, and color response. These visual discrepancies complicate matching recorded experience to the robot's current observations. Body dimensions, joint structure, and available motions also determine which actions can be executed: a person can turn sideways through a narrow passage or manipulate an object with a dexterous hand, whereas a mobile manipulator has a different footprint and gripper. Transferring experience therefore requires interpreting observations across cameras and viewpoints and grounding actions in the robot's own geometry and kinematics.

\textit{Third, navigation and manipulation are not explicitly separated in human activity.} People naturally adjust their approach and position as they get ready to interact, without marking when navigation ends and manipulation begins. For a robot, reaching the vicinity of an object does not ensure that it can manipulate it. The recording does not explicitly reveal which movements simply bring a person closer and which make the subsequent interaction possible.

To overcome these challenges, we present \textbf{MoCaP}, which turns passive human video into guidance for mobile navigation and manipulation (Fig.~\ref{fig:figure1}). MoCaP extracts spatial memory about a place together with records of where people stood when they reached for objects. During task execution, the spatial memory helps locate the requested object, and the human approach prior guides the robot to approach the object in the right way. A code agent uses this guidance together with current observations to coordinate navigation and manipulation primitives. The primitives solve and check motions for the robot's embodiment, returning execution feedback that lets the agent revise subsequent actions. %

Our contributions are:
\begin{itemize}
\item \textbf{A unified Code-as-Policy system for mobile manipulation.} MoCaP flexibly coordinates navigation and manipulation through executable code, without a predefined division into navigation and manipulation stages. The agent selects and revises actions using prior human experience, live observations, and feedback from robot-grounded primitives.
\item \textbf{Reusable spatial memory from passive human video.} We develop a fine-grained spatial memory pipeline that draws on the abundance of everyday human activity recordings to extract object relations, landmarks, and place connectivity for subsequent robot search and navigation. This video is task-independent and can be reused to different tasks.
\item \textbf{Human-guided base placement for manipulation.} We infer the direction to approach object from how a person positioned themselves before manipulation in passive human videos, then search nearby base poses under the robot's own reachability and collision constraints. A poor arrival side can leave no feasible pose within reach of local adjustment, and this guidance is designed to turn such arrivals into graspable configurations, using only video that accumulates from everyday activity.
\end{itemize}

We evaluate MoCaP on a real mobile manipulator in an indoor environment through an object-navigation task and two mobile manipulation tasks. MoCaP achieves 80\% success on each manipulation task, whereas existing methods~\cite{capx} cannot succeed at the task. To examine the contributions of video-derived guidance, we separately evaluate spatial memory for object search and human approach priors for base placement. Spatial memory increases navigation success from 70\% to 100\%, while controlled comparisons suggest that human approach priors improve manipulation readiness from unfavorable initial base configurations. 
Base placement guided by the human prior found a reachable pose in 5 of 5 trials whereas a naive proximity-based docking failed in all five.
Together, these results suggest that passive human video can help robots complete mobile manipulation tasks by guiding both object search and the approach to manipulation.

%% file: figures/figure2.tex
\begin{figure*}[t]
\centering
\includegraphics[width=\textwidth]{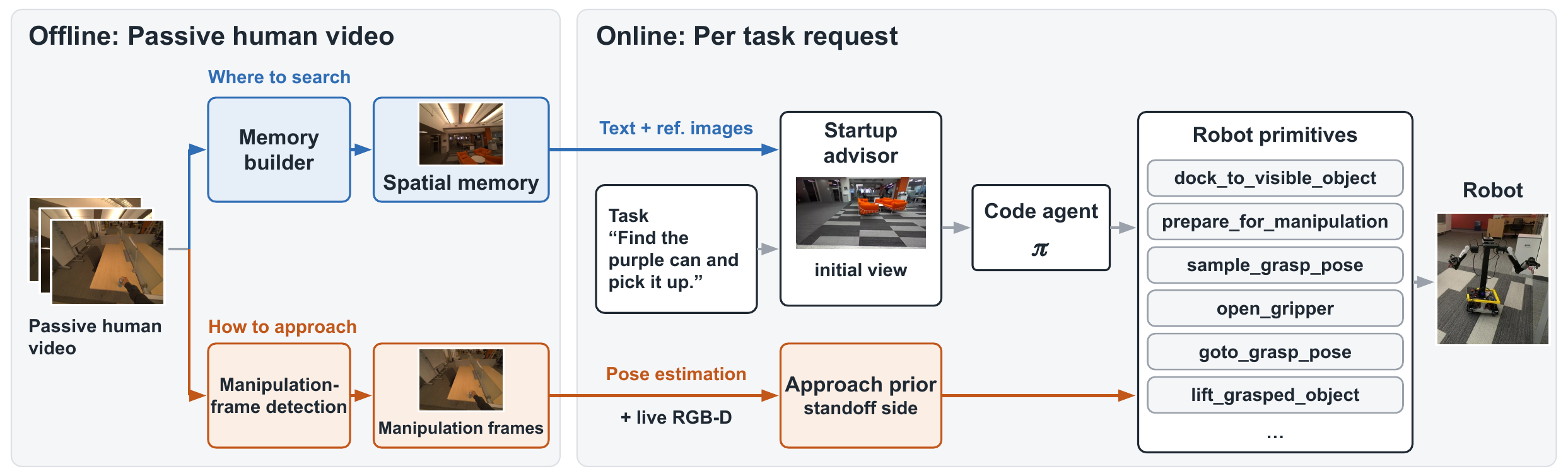}\caption{\textbf{MoCaP overview.} Passive human video provides reusable spatial memory and observations of how people prepare to manipulate. At task startup, an advisor VLM relates the spatial memory to the instruction and initial robot view. During execution, the system retrieves a recorded interaction with the target object and estimates where the person in the video stood relative to it from the frame preceding the reach.
A closed-loop code policy coordinates navigation and manipulation through planning primitives that check actions against live geometry and robot constraints, revising execution from observations and feedback.}
\label{fig:figure2}
\end{figure*}

%% file: sections/related_work.tex
\section{Related Work}
\subsection{Human Video as Scene and Interaction Experience}
Using past observations to guide future action is a central motivation for persistent scene representations. Spatial-semantic maps and scene graphs associate objects with locations, appearance, and spatial relations, supporting queries and planning beyond the robot's current
view~\cite{vlmaps,conceptgraphs,dynamem}. More recently, recorded video has provided another source of scene experience: relational video memories support later scene questions~\cite{vlkng}, while tour-conditioned navigation retrieves goal views for graph-based execution~\cite{mobilityvla} or generates route instructions for a downstream navigator~\cite{videonig}. Together, these approaches establish how earlier observations can help an agent determine where to search. For mobile manipulation, however, locating the relevant workspace is only part of the problem: the robot must also approach it from a
configuration that permits interaction.

Human activity video offers complementary information about how
interaction is prepared and performed. Prior work transfers predicted
hand trajectories into robot actions~\cite{passiveactions}, or extracts
contact locations and subsequent motion as visual affordances for robot learning~\cite{vrb}. SafeMimic adapts demonstrated actions through safety-checked search~\cite{safemimic}, and HoMMI learns hand-eye trajectories realized by whole-body control~\cite{hommi}. 
These approaches motivate using human behavior as more than a record of object locations. MoCaP focuses on a complementary cue: the position from which a person prepares to interact. It combines this cue with spatial memory from the same passive recordings, using scene experience to guide search and pre-contact positioning to guide approach. The recordings are processed before downstream requests are supplied and can support multiple tasks without a prescribed demonstration of each requested robot behavior.

\subsection{Language-Guided Mobile Manipulation}
Language-guided robot systems connect high-level requests to executable behavior through perception and action interfaces. Skill-based approaches use language or vision-language models to select and parameterize robot capabilities~\cite{saycan,bumble}; Code as Policies extends this interface to executable programs that compose perception and control APIs into executable programs~\cite{cap}. Incorporating execution feedback enables these systems to revise plans and recover from failures~\cite{innermonologue,come}, while ASPIRE turns validated program repairs into reusable skills~\cite{aspire}. For mobile manipulation, integrated systems bring object search, navigation, and manipulation into a common execution pipeline~\cite{homerobot,okrobot}.

CaP-X provides a closely related foundation for composing perception and control primitives, with its real-robot evaluation centered on tabletop manipulation~\cite{capx}. Its mobile manipulator interface selects base poses by geometric heuristics and relies on the simulator's planner to reject infeasible ones. MoCaP adopts the CaP-X program interface but grounds base motion in the real robot: docking goals are validated against the online costmap and executed by a classical navigation stack~\cite{nav2}, and arrival is verified with fresh perception before manipulation proceeds. Moving to mobile manipulation also introduces a coupled problem beyond collision-free motion: the robot must decide where to search and from which side to approach for manipulation. A code-based control interface enables these decisions but does not itself supply the scene and interaction experience needed to guide them, failing the tasks. MoCaP draws this guidance from passive human video, using spatial memory to guide search and pre-contact positioning to guide approach, while robot-specific primitives determine executable motions.

\subsection{Base Placement for Mobile Manipulation}
Whether a mobile manipulator can execute a manipulation task heavily depends on where it stops.
Classical methods precompute a voxelized map of the arm's reachable workspace~\cite{zacharias2007capturing}
and invert it to obtain the base poses from which a target is reachable~\cite{vahrenkamp2013orm}. 
These maps depend only on kinematics, so they are nearly symmetric around the target and cannot tell which side of a table is free or which way a door opens.
Learning-based methods recover that context from successful rollouts~\cite{n2m}, reinforcement learning with reachability prior~\cite{jauhri2022reachability}, at the cost of robot trials in each new environment.
Welschehold et al.~\cite{welschehold2017learning} initialized base poses from observed human torso poses, yet their demonstrations are specific to target task. 
We instead use passive egocentric video, detecting the transition from walking to reaching and using the recorded human standing position as a good prior to order the search over base poses.

%% file: sections/method.tex
\section{Method}
\subsection{Overview}
MoCaP takes a language instruction $q$, a prior human RGB recording $V=\{I^h_\tau\}$, and live robot observations $o_t=(I_t,D_t,x_t,z_t)$ containing RGB, depth, base pose, and joint state. Camera calibration, robot kinematics, and online local costmaps are available; the recording supplies no metric camera poses or target coordinates. All learned components remain frozen.

Video processing produces reusable spatial memory and records of human interactions. At task startup, an advisor uses the memory, instruction, and initial robot view to provide guidance
to a code agent. During execution, the code agent generates short Python programs that invoke robot-grounded primitives and revises subsequent programs using fresh observations and execution feedback. The human preparation prior guides docking, while the primitives resolve motions using live geometry and robot constraints (Fig.~\ref{fig:figure2}).

\subsection{Offline Spatial and Interactive Memory Construction}
An offline builder organizes the recording into text $m$ and timestamped reference frames $E$:
\begin{equation}
 M=(m,E)=\mathcal B(V),\qquad E=\{(\tau_k,I^h_{\tau_k})\}_{k=1}^{K}.
\end{equation}
The builder uses GPT-5.6~\cite{gpt56} to generate scene descriptions $m$ that cite relevant video timestamps, then extracts the corresponding reference frames $E$. Together, the text and frames form the spatial memory $M$. The text retains places, connections, landmarks, and object descriptions with support surfaces and relative locations. The builder links small-object observations to their surrounding regions and records transition cues: where to turn, which landmark should become visible next, and how to recognize a revisited place. Ambiguous identities and inferred connections remain marked as uncertain. Up to 32 frames are extracted at timestamps cited by the memory, providing visual references for these descriptions.

Given the instruction and initial robot image, a single startup advisor computes $a_0=\mathcal A(M,q,I_0)$. It receives the memory text and reference images with historical and live views explicitly distinguished. Its output describes the robot's likely location, two to four navigation subgoals with progress cues, and the target's appearance or support surface. This guidance is added to the code agent's initial context and remains available through conversation history. The advisor is called once per task; the agent uses subsequent observations to revise how it follows the guidance.

A separate event set $\mathcal H$ records human interactions and the views preceding them. Mean grayscale frame differences identify low-motion windows, sampled at 2~Hz for VLM labeling. For each interaction, the model identifies the object, hand, last frame before reaching, and first contact. We save the ready frame, a view 2~s earlier, and the contact frame. The ready view captures the person's position before arm motion changes the view or occludes the target; the preceding view provides additional scene context for geometric association. Both memory and event extraction precede downstream instructions.

\subsection{Code-as-Policy and Robot-Grounded Primitives}
We implement closed-loop control through a Code-as-Policy module, in which a vision-language code agent generates Python programs that invoke robot-grounded primitives.
We denote the agent's code-generation policy by $\pi$.
Given the instruction $q$, startup guidance $a_0$,
current visual observation and state summaries $\tilde o_t$,
and conversation history $h_t$, the agent generates a
program $c_t$, which the runtime executes:
\begin{equation}
 c_t=\pi(q,a_0,\tilde o_t,h_t),\
 (r_t,o_{t+1})=\operatorname{ExecuteObserve}(c_t).
\end{equation}
Execution feedback $r_t$ and a fresh observation inform the next code-generation step. Failed action calls interrupt the remaining program, allowing the agent to revise its plan. History persists in the conversation. A typical manipulation sequence docks to a visible object, prepares the selected arm, samples a checked grasp, and requests its execution. If preparation fails, the subsequent grasp call is skipped, and the agent can change its approach using the returned diagnostics.

SAM~3~\cite{sam3} masks and live depth locate named targets. The docking primitive converts a visible target into a ground-plane standoff goal facing the object; Nav2~\cite{nav2} handles navigation, and fresh perception verifies arrival. Bounded turning, forward/backward motion, and lateral translation support search and recovery when no docking target is available. These primitives ground video-derived guidance in the robot's own viewpoint and motion capabilities.

For manipulation, GraspGen-X~\cite{graspgenx} proposes grasps for the segmented target. Calibrated transforms convert model grasps to tool center point (TCP) goals, and open-gripper swept-volume checks reject approaches obstructed by observed non-target geometry. The following stages determine where the robot should stand and which grasps it can execute.

\subsection{Approaching Object under Guidance from Human Prior}
\label{sec:human_prior}

At docking, we retrieve the first event in $\mathcal H$ that shares a non-article word with the target description. Pretrained VGGT~\cite{vggt} takes the robot RGB image and the event's preceding and ready views to estimate camera intrinsics and relative poses. We align the human camera with the robot camera using live depth, obtaining the human camera position $C_h$ in the robot frame:
\begin{equation}
\alpha=\operatorname{median}_{p\in\Omega}
\frac{D_t(p)}{\widehat D_t(p)},\qquad
C_h=-\alpha R_h^\top t_h,
\label{eq:human_scale}
\end{equation}
where $\Omega$ contains valid depth pixels and $\alpha$ provides the metric scale. We project $C_h$ and its optical axis onto the ground plane to obtain the human approach position $p_h$ and heading $\theta_h$. We then approximate the human stance as
\begin{equation}
p_s=p_h-\delta e(\theta_h),
\qquad
e(\theta)=(\cos\theta,\sin\theta)^\top,
\end{equation}
and derive the approach bearing toward the target position $p_o$ as
\begin{equation}
\beta_h=
\begin{cases}
    \theta_h+\pi, & \|p_h-p_o\|\leq\epsilon,\\
    \operatorname{atan2}((p_s-p_o)_y,(p_s-p_o)_x), & \text{otherwise}.
\end{cases}
\label{eq:human_bearing}
\end{equation}
When the human is already close to the object, the heading provides a more stable estimate than the relative position.

The resulting bearing provides a human-derived approach prior rather than a fixed robot goal. We transfer $\beta_h$ to the robot while choosing a robot-specific standoff distance $\rho$ and checking reachability through local grasp and IK constraints. The desired camera position is $p_o+\rho e(\beta_h)$, from which the corresponding chassis goal is computed using the camera-to-base transform. If the resulting navigation goal is infeasible, we retry nearby approach directions before falling back to the original arrival direction. The grasping arm is selected according to the target's side relative to the robot, with the human hand used to resolve ambiguous cases. Thus, human observations determine \emph{where and from which direction to approach}, while robot-specific geometry determines the executable configuration.

\subsection{Manipulation Preparation and Grasp Execution}
\label{sec:preparation}

The manipulation preparation primitive searches nearby base configurations before moving. 
For each candidate base pose, we transform the retrieved grasp poses into the corresponding arm-base frame and use gripper collision and IK checks to efficiently filter infeasible configurations before invoking whole-arm trajectory planning. 
We require the target to be visible within 0.8~m and evaluate a local grid of planar translation and yaw adjustments around the current base pose. 
The search includes the zero adjustment, allowing an already suitable pose to proceed directly to grasping.

Let $T_{XY}$ denote the transformation from frame $Y$ to frame $X$. 
Let $C_0$ and $C_u$ denote the current and candidate camera frames, and let $H(u)=T_{C_0C_u}$ be the induced camera motion. 
For a grasp pose $G_j=T_{C_0G_j}$, the corresponding candidate TCP goal is
\begin{equation}
 T_{A,j}(u)=T_{AC}H(u)^{-1}G_jT_{GE},
 \label{eq:virtual_grasp}
\end{equation}
where $T_{AC}$ is the fixed camera-to-arm-base calibration for the selected arm and $T_{GE}$ expresses the TCP frame $E$ in the grasp-model frame $G$. 
Thus, the grasp is first expressed in the candidate camera frame and then transformed into the arm-base frame.

We shortlist candidate base poses using workspace constraints and a lightweight geometric score, then retain poses for which at least one grasp passes gripper collision and nominal IK checks. 
Among feasible candidates, we prefer small base adjustments while favoring configurations with multiple and robustly feasible grasps. 
The resulting grasp set and base configuration are stored together for subsequent execution.

The selected adjustment is executed as an absolute planar base goal. 
After motion, grasp feasibility is re-evaluated using fresh observations because controller error can invalidate the virtual prediction. 
The retained alternatives are then passed to cuRobo~\cite{curobo}, which plans whole-arm motion through a pre-grasp pose and constrained approach using the measured robot state and fresh scene geometry. 
Planning or execution failure returns control to the agent.

Spatial memory guides where to search, while the human approach prior guides the initial base placement. 
The code agent combines these priors with live observations and execution feedback, while robot-specific feasibility checks determine the motions that can be executed.

%% file: sections/experiments.tex
\raggedbottom
\section{Experiments}
\label{sec:experiments}
\input{figures/figure3}
\input{figures/figure4}
Our experiments evaluate MoCaP on a real mobile manipulator in open-vocabulary manipulation tasks. We ask three questions: (i) does MoCaP complete tasks that existing modular systems fail to complete? (ii) does the scene memory built from human video reduce the search required to locate an object? (iii) does the human approach prior for base placement increase the chance of reaching a manipulation-ready base pose?

\subsection{Experimental Setup}
Experiments use the YOR platform~\cite{yorplatform} shown in Fig.~\ref{fig:figure3}. The robot comprises an omnidirectional base, a vertical lift, two AgileX Nero arms with grippers, and a ZED2i stereo RGB-D camera. A Jetson Orin NX hosts the agent runtime and perception/planning services, with Raspberry Pi services controlling the base and arms. The code agent calls GPT-5.6~\cite{gpt56} with the reasoning effort (thinking level) set to medium. Figure~\ref{fig:figure4} shows the indoor environment, navigation to a trash bin, and two mobile manipulation tasks.

Human activity videos are recorded using Aria Gen~2 glasses~\cite{aria_gen2_egocentric_ai_2025}. We use only the recorded egocentric RGB video; no other data provided by the glasses are used. GPT-5.6 performs offline video understanding and spatial-memory construction before downstream instructions are supplied, and the same memory is reused across tasks. We evaluate one object-navigation task, navigating to a blue trash bin, and two mobile manipulation tasks, grasping a cardboard box on a white storage cabinet and grasping a can on a table.

Comparisons use matched tasks, object placements, and robot starts, with scene resets between trials. Model weights, prompts, perception backends, and primitive settings remain fixed except for the interface under study. Complete-task outcomes are judged from recorded physical execution. We report successes/trials and success rate (SR), with unsuccessful episodes retained in the denominator. For navigation, we additionally report the numbers of episodes completed successfully within 180 and 300 seconds to evaluate how quickly the robot could navigate.

\subsection{Complete Mobile Manipulation}
We evaluate MoCaP on two tasks: grasping a cardboard box and grasping a can, each illustrated in Fig.~\ref{fig:figure4}\subref{fig:grasp_box} and \subref{fig:grasp_can}. The robot approaches the target, prepares a grasping configuration, and acquires the object. The box task starts with the target in view, providing a simpler setting for evaluating approach and grasping. The can is outside the robot's initial view, so the robot must first locate it before preparing to grasp it. Success requires grasping the requested object and keeping it securely held.

We compare MoCaP with CaP-X~\cite{capx}, an agentic framework for robotic manipulation. We run it through its mobile manipulator (R1Pro) API, which exposes SAM3 and Contact-GraspNet~\cite{contactgraspnet} perception, base navigation and in-place search, IK, and grasping, ported to our robot with the same task instructions, policy model, time limit, and program budget as MoCaP. Base goals are executed by the platform's base controller, and the operator stops the robot before a collision, which counts as a failure.

CaP-X selects a standing pose with a geometric heuristic and relies on the simulator's planner to reject infeasible goals. On the real robot, an infeasible goal is executed as given. To separate this execution gap from the choice of where to go, we also evaluate CaP-X+, which keeps the CaP-X API, prompt, and manipulation unchanged but executes base goals with collision-aware navigation. Each goal is checked against the online costmap with the full robot footprint, moved to the nearest collision-free pose if needed, and reached with Nav2~\cite{nav2}.

Table~\ref{tab:main} reports the number of trials in which the robot reached the target and attempted a grasp, and the number of successful trials. MoCaP succeeds in 16/20 box trials and 8/10 can trials (80\% each). Neither baseline succeeds. On the box, collision-aware navigation increases grasp attempts from 2/20 for CaP-X to 8/20 for CaP-X+, but it does not make the heuristic standing pose reachable. When the goal falls behind or inside the cabinet, the nearest collision-free pose is often out of arm's reach, and five of the eight CaP-X+ attempts started 0.74--1.64\,m from the box. Of the other three attempts, two ended with the arm striking the furniture, because the CaP-X grasping primitive moves the arm to an inverse kinematics solution by joint-space interpolation without collision checking. MoCaP instead plans whole-arm motion with cuRobo~\cite{curobo} against the observed scene geometry (Sec.~\ref{sec:preparation}). In the third, the grasp target lay outside the arm's workspace. On the can, which starts out of view, both baselines search only by rotating in place and never localized the target. No CaP-X trial localized the can within the full time limit, so we ended a CaP-X+ trial after two complete rotations without a detection or base motion (8 of 10 trials).

\begin{table}[t]
\centering
\caption{Mobile manipulation results.}
\label{tab:main}
\begin{tabular}{@{}llcc@{}}
\toprule
Task & Method & Grasp attempted & Success\\
\midrule
Grasp box & \textbf{MoCaP (ours)} & \textbf{19/20} & \textbf{16/20 (80\%)}\\
              & CaP-X+ & 8/20 & 0/20 (0\%)\\
              & CaP-X~\cite{capx} & 2/20 & 0/20 (0\%)\\
\midrule
Grasp can & \textbf{MoCaP (ours)} & \textbf{10/10} & \textbf{8/10 (80\%)}\\
              & CaP-X+ & 0/10 & 0/10 (0\%)\\
              & CaP-X~\cite{capx} & 0/10 & 0/10 (0\%)\\
\bottomrule
\end{tabular}
\end{table}

\subsection{Spatial Memory and Object Navigation}
We evaluate the navigation ability of MoCaP with a task where the robot needs to navigate to a blue trash bin from a place where it can't see the target, as illustrated in Fig.~\ref{fig:figure4}\subref{fig:navigate_trash_bin}.

\begin{table}[!t]
\centering
\caption{Blue-trash-bin navigation from the kitchen.}
\label{tab:navigation}
\begin{tabular}{@{}lccc@{}}
\toprule
Method & Successes/trials & \shortstack{Success within\\180\,s} & \shortstack{Success within\\300\,s}\\
\midrule
\textbf{MoCaP (ours)} & \textbf{10/10} & \textbf{6/10} & \textbf{9/10}\\
MoCaP (no prior) & 7/10 & 4/10 & 6/10\\
\midrule
ApexNav~\cite{apexnav} & 2/10 & 1/10 & 2/10\\
CoW~\cite{cow} & 0/10 & 0/10 & 0/10\\
\bottomrule
\end{tabular}
\end{table}

Table~\ref{tab:navigation} compares MoCaP, a no-prior variant, and two external baselines on navigation to the blue trash bin from the kitchen start. CoW~\cite{cow} selects frontiers geometrically and uses open-vocabulary localization only to decide when the target is found. ApexNav~\cite{apexnav} scores frontiers by semantic reasoning and falls back to geometry when semantic cues are weak. We chose these two because they span the range of exploration strategies available without an offline prior. Neither can consume such a prior, so we compare them against the no-prior variant, which operates under the same assumption, and reserve the comparison with full MoCaP for measuring the prior itself. We run ten episodes per method. An episode succeeds if the robot's base stops within 1,m of the blue trash bin and the bin is visibly identified in the terminal robot view. We also report success within fixed time budgets because the prior is intended to improve efficiency in addition to success rate.

MoCaP succeeds in 10/10 episodes, compared with 7/10 for the no-prior variant, 2/10 for ApexNav, and 0/10 for CoW. Within 180 seconds the respective counts are 6, 4, 1, and 0, and within 300 seconds they are 9, 6, 2, and 0. The gap between MoCaP and the no-prior variant is consistent with a benefit from prior guidance in this setting.The external baselines fail in characteristic ways. CoW consistently stops at a white trash bin in the kitchen that carries a blue notice on its side. Its localizer assigns high confidence to this distractor, and since CoW terminates once confidence exceeds the threshold, it never leaves the starting room. ApexNav fails in two ways. In some episodes it detects the blue bin from a distance but continues exploring instead of committing to it, so the remaining navigation exceeds the time budget. In the others it settles on a small black shredder in an adjacent room and declares the goal reached.

\subsection{Effect of Human Approach Prior in Base Placement}
\label{sec:prior_experiment}
As the robot approaches an object, its direction of arrival shapes the arm's reach and the local adjustments needed for grasping. The human's position before interaction provides a reference for this approach. We examine how this guidance affects physical grasping within a fixed time budget and the preparation effort accumulated during execution.

\input{figures/figure5}

The robot starts from A, B, or C in Fig.~\ref{fig:figure5}; the human interaction occurs on the C side. From each matched start, we compare with and without the human approach prior. Each trial executes a fixed sequence of docking followed by local preparation, without the code agent, so neither condition can recover from a failed approach or retry from a different direction. Both conditions share the same perception, docking primitive, local search, and grasp planner. Only the approach direction supplied to docking changes: with the prior it follows the recorded human stance, and without it the robot retains its arrival direction.

Success denotes readiness, defined as the existence of a base pose from which at least one proposed grasp passes the gripper collision checks and nominal IK. We report five trials per start and condition and classify each failure by the stage at which it occurred.

\begin{table}[t]
\centering
\caption{Manipulation readiness (fifteen trials per start and condition).}
\label{tab:human_prior}
\setlength{\tabcolsep}{6pt}
\begin{tabular}{@{}lccc@{}}
\toprule
& \multicolumn{2}{c}{Success rate (\%)} & \\
\cmidrule(lr){2-3}
Start & Without prior & With prior & $\Delta$ \\
\midrule
A   & 0.0  & \textbf{93.3} & $+93.3$ \\
B   & 26.7 & \textbf{80.0}  & $+53.3$ \\
C   & \textbf{93.3} & 80.0  & $-13.3$ \\
\midrule
All & 40.0 \scriptsize{(18/45)} & \textbf{84.4} \scriptsize{(38/45)} & $+44.4$ \\
\bottomrule
\end{tabular}
\end{table}

Table~\ref{tab:human_prior} reports the results.
With the prior, readiness is reached in 38/45 trials, compared with 18/45 without it. The effect is concentrated at A, where the prior changes 0/15 to 14/15, while C shows no advantage. This follows the layout in Fig.~\ref{fig:figure5}: the recorded interaction occurs on the C side, so the arrival direction without the prior already coincides with the human stance there, and A is the start for which the two most disagree. The pooled comparison is therefore not the quantity of interest, and at this sample size it is not significant.

%% file: figures/figure3.tex
\begin{figure}[t]
\centering
\includegraphics[width=.7\linewidth]{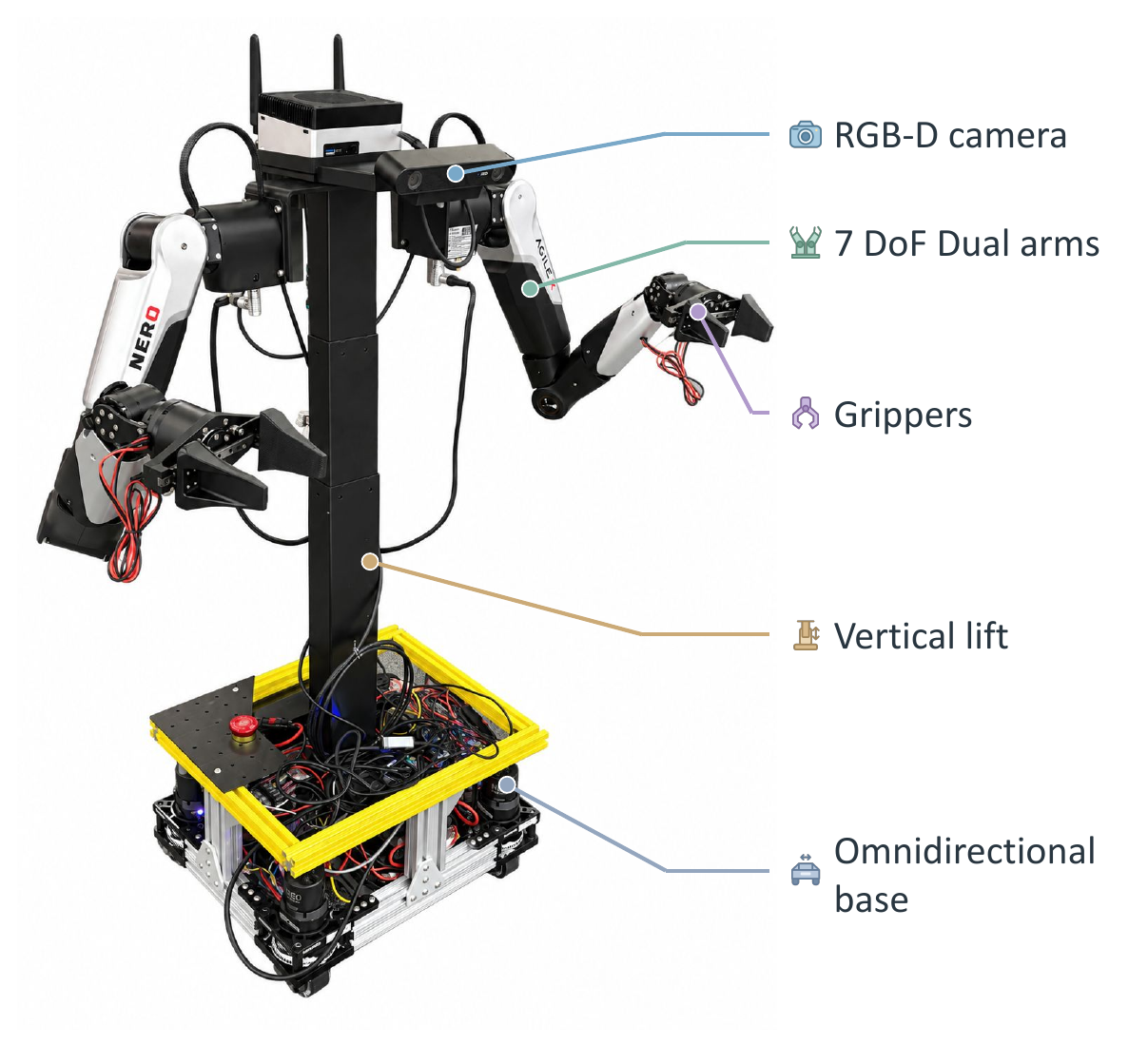}
\caption{YOR robot platform used in our experiments, equipped with an RGB-D camera, dual arms with grippers, a vertical lift, and an omnidirectional base.}
\label{fig:figure3}
\end{figure}

%% file: figures/figure4.tex
\begin{figure*}[t]
\centering
\captionsetup[subfloat]{position=top,justification=raggedright,singlelinecheck=false}
\subfloat[Grasp cardboard box]{%
  \label{fig:grasp_box}%
  \includegraphics[width=0.9\textwidth]{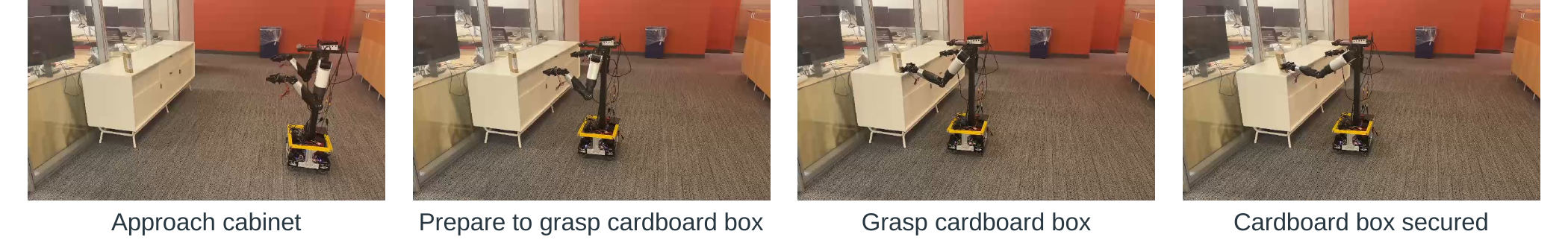}}\par
\subfloat[Grasp can]{%
  \label{fig:grasp_can}%
  \includegraphics[width=0.9\textwidth]{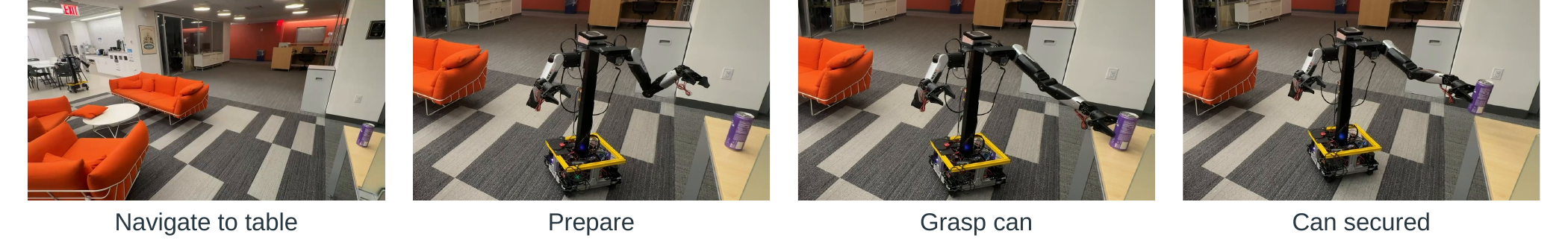}}\par
\subfloat[Navigate to trash bin]{%
  \label{fig:navigate_trash_bin}%
  \includegraphics[width=0.9\textwidth]{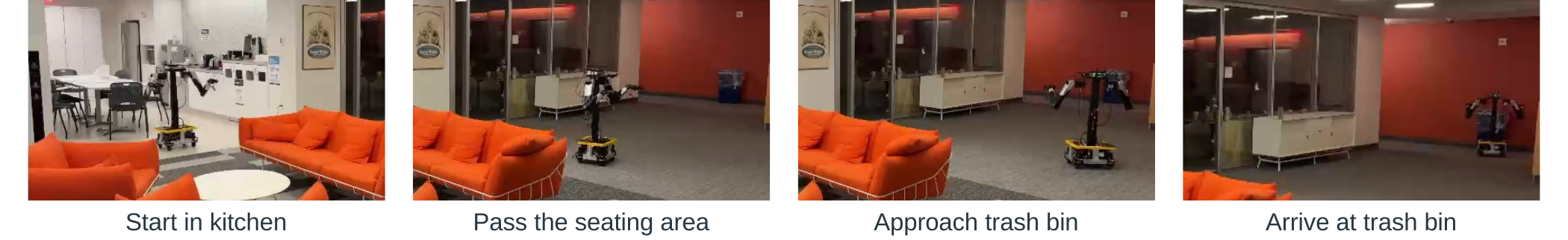}}
\caption{Example executions of \protect\subref{fig:grasp_box} grasping a cardboard box, \protect\subref{fig:grasp_can} grasping a can, and \protect\subref{fig:navigate_trash_bin} navigating to a blue trash bin. Keyframes in each row follow one continuous execution from left to right.}
\label{fig:figure4}
\end{figure*}

%% file: figures/figure5.tex
\begin{figure}[t]
\centering
\begin{tikzpicture}[x=.82cm,y=.82cm,font=\small,
  base/.style={draw=blue!50!black,fill=blue!6,rounded corners=2pt,
               minimum width=.95cm,minimum height=.60cm,inner sep=1pt},
  humanbase/.style={base,draw=teal!65!black,fill=teal!8},
  look/.style={->,>=stealth,densely dotted,black!40,
               shorten >=4pt,shorten <=2pt}]

\filldraw[fill=black!4,draw=black!35,rounded corners=2pt,line width=.4pt]
  (2.3,0) rectangle (6.1,2.3);
\node[text=black!45,font=\footnotesize] at (5.35,1.95) {Desk};

\filldraw[fill=violet!70,draw=violet!80!black,line width=.4pt]
  (2.85,.62) coordinate (can) circle (.11);
\node[right=3pt,text=violet!75!black,font=\footnotesize] at (can) {Can};

\node[base]      (A) at (3.00,3.15) {A};
\node[base]      (B) at (1.05,2.25) {B};
\node[humanbase] (C) at (1.05,0.62) {C};

\draw[look] (A) -- (can);
\draw[look] (B) -- (can);
\draw[look] (C) -- (can);

\node[below=2pt,text=teal!65!black,font=\footnotesize] at (C.south) {Human side};
\end{tikzpicture}
\caption{Schematic layout of the base placement study. A, B, and C denote the three starting configurations around the desk, and the dotted lines indicate the direction to the target can. The recorded human interaction occurs on the C side.}
\label{fig:figure5}
\end{figure}
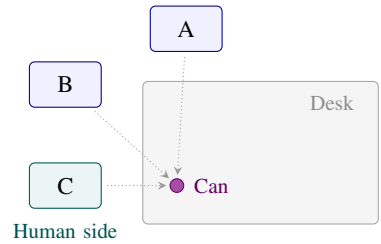

%% file: sections/conclusion.tex
\section{Limitations and Conclusion}
We presented MoCaP, which turns task-independent passive human video into guidance for open-vocabulary mobile manipulation. The video supplies where to search, from a spatial memory of the recorded environment, and from which side to approach, from where a person stood before reaching for an object. Combined with live perception and robot-specific planning, this guidance let a real mobile manipulator complete tasks that a tabletop coding agent could not.

MoCaP has several limitations. First, our evaluation is limited to one environment, one video recording, and three target objects with five to ten trials per condition. The results therefore establish feasibility rather than generality. Whether memory and approach priors transfer across rooms, recorders, and objects absent from the recording remains open. Second, the system design itself constrains performance. A fixed startup summary can omit relevant objects or become stale. Local base search depends on visible geometry and the grasp model and does not optimize future visibility. Partial depth observations and imperfect footprint models limit collision checks. Manipulation is limited to grasping and lifting. Future work could evaluate the priors across multiple homes and recordings, and use human video to guide richer skills involving sustained contact, precise placement, or coordinated use of both arms.